\documentclass{article}
\usepackage{arxiv}

\usepackage[utf8]{inputenc} 
\usepackage[T1]{fontenc}    
\usepackage{hyperref}       
\usepackage{url}            
\usepackage{booktabs}       
\usepackage{amsfonts}       
\usepackage{nicefrac}       
\usepackage{microtype}      
\usepackage{lipsum}
\usepackage{graphicx}
\graphicspath{ {./images/} }

\usepackage{microtype}
\usepackage{graphicx}
\usepackage{subfigure}
\usepackage{natbib}

\usepackage{amsmath}
\usepackage{amssymb}
\usepackage{mathtools}
\usepackage{amsthm}

\usepackage[capitalize,noabbrev]{cleveref}

\theoremstyle{plain}
\newtheorem{theorem}{Theorem}[section]

\theoremstyle{definition}
\newtheorem{definition}[theorem]{Definition}

\theoremstyle{remark}
\newtheorem{remark}[theorem]{Remark}

\usepackage{threeparttable, array, float} 
\usepackage{microtype}
\usepackage{graphicx}
\usepackage{algorithm}
\usepackage{algorithmic}
\usepackage{diagbox}
\usepackage{booktabs} 
\usepackage{amsmath}
\usepackage{amsthm}
\usepackage{color, colortbl}
\usepackage{hyperref}
\usepackage[capitalize]{cleveref}
\usepackage{nicefrac} 
\usepackage{amsmath,amsfonts,bm}

\definecolor{dark-red}{rgb}{0.4, 0.15, 0.15}
\definecolor{dark-blue}{rgb}{0.15, 0.15, 0.4}
\definecolor{medium-red}{rgb}{0.5, 0, 0}
\definecolor{medium-blue}{rgb}{0, 0, 0.5}
\definecolor{light-red}{rgb}{0.7, 0, 0}
\definecolor{light-blue}{rgb}{0, 0, 0.7}

\title{LLM Cache Bandit Revisited: Addressing Query Heterogeneity for Cost-Effective LLM Inference}

\author{
Hantao Yang \\
  University of Science and Technology of China\\
  \texttt{yanghantao@mail.ustc.edu.cn} \\
   \And
Hong Xie\\
  University of Science and Technology of China\\
  \texttt{xiehong2018@foxmail.com} \\
  \And
Defu Lian\\
  University of Science and Technology of China\\
  \texttt{liandefu@ustc.edu.cn} \\
  \And
Enhong Chen\\
  University of Science and Technology of China\\
  \texttt{cheneh@ustc.edu.cn} \\
}

\begin{document}
\maketitle
\begin{abstract} 
This paper revisits the LLM cache bandit problem, with a special focus on addressing the query heterogeneity for cost-effective LLM inference. Previous works often assume uniform query sizes. Heterogeneous query sizes introduce a combinatorial structure for cache selection, making the cache replacement process more computationally and statistically challenging.  We treat optimal cache selection as a knapsack problem and employ an accumulation-based strategy to effectively balance computational overhead and cache updates. In theoretical analysis, we prove that the regret of our algorithm achieves an $O(\sqrt{MNT})$ bound, improving the coefficient of $\sqrt{MN}$ compared to the $O(MN\sqrt{T})$ result in Berkeley, where $N$ is the total number of queries and $M$ is the cache size. Additionally, we also provide a problem-dependent bound, which was absent in previous works. The experiment rely on real-world data show that our algorithm reduces the total cost by approximately 12\%.
\end{abstract}

\section{Introduction}

Prominent Large Language Models (LLMs), such as GPT \cite{GPT}, Gemini \cite{gemini}, and DeepSeek \cite{DeepSeek}, have demonstrated exceptional capabilities, driving innovations in AI-powered systems. However, as Large Language Models (LLMs) significantly enhance the capabilities of AI systems, the increasing volume of query processing requests presents challenges for cost-effective inference, particularly due to repetitive queries that lead to unnecessary resource consumption and increased costs. There are several cost-effective solutions for managing LLM applications. \cite{frugalgpt} explored methods like prompt engineering to optimize queries before sending them to public LLMs. In the real-world marketplace of LLMs, it is inevitable for online large model systems to handle some repetitive queries \cite{frugalgpt}, which leads to higher inference costs and longer latency compared to traditional models \cite{bang-2023-gptcache}. As the scale of LLMs grows larger, the cost of repeated queries becomes increasingly significant. Developing methods to reduce these redundant costs and resource waste would be highly beneficial for the efficient deployment of large models.

Cache techniques are an effective way to reduce repeated queries in LLMs, minimizing the impact of redundant inquiries. Recently, caching strategies tailored for LLMs have emerged. H2o \cite{zhang2023h2oheavyhitteroracleefficient} optimize internal data flow in LLMs to reduce delays and resource consumption during inference. GPTcache \cite{bang-2023-gptcache}  is known for its simplicity and flexibility, SCALM \cite{SCALM} improves cache efficiency by leveraging the semantic understanding of queries. \cite{Berkeley} considered a bandit setting for online cache updates based on estimated values. These works have successfully integrated caching strategies into the deployment of large models, but most of them do not consider the settings with variable size query.

Incorporating cache into large language models (LLMs) poses challenges due to the variability and unpredictability of query costs. To address this issue, \cite{Berkeley} suggests using both expected cost and popularity as criteria for cache selection in their online algorithm. While this approach works well when query sizes are fixed, it lacks solutions for cases where query sizes vary. In such scenarios, the length of the query must also be considered, making the number of queries that can be stored in the cache no longer constant, a point that \cite{Berkeley} briefly discusses but leaves open for further improvement.

To tackle this open problem, we revisit the framework of \cite{Berkeley} to accommodate heterogeneous query sizes in online query selection for LLM systems. By applying the model to this more generalized scenario, we address not only the feedback scarcity issue inherent to traditional bandit settings \cite{Berkeley} but also additional challenges: (1) Heterogeneous query sizes transform the cache replacement process into a combinatorial optimization problem. To accommodate a new query, the cache may need to evict multiple existing queries. This multiple-query replacement strategy requires careful selection of which queries to evict to maximize the cache hit rate while adhering to size constraints. One possible approach is to replace the entire cache at once, treating the selected cache as the solution to a combinatorial optimization problem. However, this introduces significant computational complexity. (2) Unlike the CMAB model \cite{CUCB}, where action selection can occur globally across all queries, online cache updates are restricted to queries currently in the cache or the incoming query. This constraint makes it difficult to directly apply traditional combinatorial optimization solutions to cache replacement in dynamic settings.

In this paper, we propose an online algorithm \textbf{Variable Size Online Cache Bandits}, an extension variant of \cite{Berkeley} to solve these issue. Our main contributions are:

1) \textit{Cache recommendation via the knapsack problem}: We design an \texttt{Oracle} that utilizes a knapsack problem solver to guide the cache replacement process. This approach extends beyond single-query replacement to support more sophisticated strategies. However, solving the knapsack problem incurs computational overhead, creating a trade-off between computational cost and cache update efficiency. Calling \texttt{Oracle} too frequently leads to excessive overhead, while infrequent updates hinder effective caching. Balancing these trade-offs is essentially the key consideration for our algorithm.

2) \textit{Accumulation-based \texttt{Oracle} invoking policy}: To address this \texttt{Oracle} trade-off, we employ an accumulation-based strategy for invoking \texttt{Oracle}. Instead of calling \texttt{Oracle} for every update, we delay its invocation until the query has accumulated sufficient information about its cost or sampling probability. This controlled invocation reduces computational overhead while maintaining efficiency. We theoretically demonstrate that the number of \texttt{Oracle} calls under this policy scales logarithmically with the number of rounds, ensuring computational efficiency.

3) \textit{Recommend and wait strategy}: 
To adapt to online cache selection, we propose a recommend-and-wait strategy. This approach introduces a recommended cache that temporarily stores the optimal cache configuration identified by the knapsack problem. Instead of immediately applying the recommendation, the strategy strategically "waits" for the relevant query to arrive. Upon \texttt{Oracle} invocation, the recommended cache is provided as a guideline for the policy. In the subsequent rounds, the algorithm incrementally updates the current cache based on the recommended cache

In summary, the proposed algorithm addresses the combinatorial optimization challenges introduced by heterogeneous query sizes and is well-suited for scenarios where cache selection is constrained in an online setting. Furthermore, our algorithm effectively manages the trade-off between additional computational overhead and cache updates.

\noindent\textbf{Theoretical Analysis.} 
The heterogeneous query sizes have introduced additional challenges in analyzing the cache selection process, turning it into a combinatorial optimization problem. Meanwhile, the online update setting imposes additional constraints during cache updates. Our main theoretical contributions are as follows: 

1) \textit{Regret bound and \texttt{Oracle} calls:} We prove that our algorithm achieves $\widetilde{O}(\sqrt{MNT})$ regret bound with $O(N\log T)$ Oracle calls.

2) \textit{Generalization to heterogeneous query sizes:} The case of equal-length queries, as studied in previous works, is shown to be a special case of our approach.  In particular, \cite{Berkeley} provided a regret bound of  $\widetilde{O}(MN\sqrt{T})$ and establish a lower bound of $\Omega(\sqrt{T})$. Our algorithm extends this framework to handle heterogeneous query sizes while improving the analysis by reducing the coefficient of $\sqrt{MN}$. Furthermore, our regret bound matches the lower bound in terms of $T$, ensuring tight performance guarantees. 

3) \textit{Problem-dependent regret bound:} In addition to the general regret bound, we provide a problem-dependent regret bound of $\widetilde{O}\left(\sum_{q\in \mathcal{M}^*}\frac{P(q)MN}{\Delta_q}\right)$.
This result highlights the dependency of regret on query-specific parameters, offering deeper insights into the performance of the algorithm under varying query distributions.

\begin{table*}[!htb]
 \caption{Summary of the main results}
		\label{tab:result}	
	\centering
	\resizebox{1\columnwidth}{!}{
	\centering
	\begin{threeparttable}
	\begin{tabular}{|ccccc|}
 \hline
\textbf{Previously work}&\textbf{Heterogeneous query size} &\textbf{Exact \texttt{Oracle}?} & \textbf{Problem-dependent regret} & \textbf{Problem-independent regret}\\
  \hline    
       \textbf{\cite{Berkeley}}& $\times$ & $\times$ & $\times$  & $\widetilde{O}\left(MN\sqrt{T}\right)$\\
    \hline
    \textbf{Our Work}&\textbf{Heterogeneous query size} &\textbf{Exact \texttt{Oracle}?} & \textbf{Problem-dependent regret} & \textbf{Problem-independent regret}\\
    \hline
   \textbf{VSOCB (\cref{alg:main})}&$\surd$ &	$\surd$&$\widetilde{O}\left(\sum_{q\in\mathcal{M}^*}\frac{P(q)MN}{\Delta_q}\right)$ (\cref{thm:main_theorem})& $\widetilde{O}\left(\sqrt{MNT} \right)$ (\cref{thm:main_theorem})\\
   \textbf{VSOCB-APX}&$\surd$ &$\times$ & $\widetilde{O}\left(\sum_{q\in\mathcal{Q}}\frac{P(q)N (N-l_{min})}{\Delta_q^c}\right)$ (\cref{thm:approximate_theorem})&$\widetilde{O}\left(\sqrt{N(N-l_{min})T}\right)$ (\cref{thm:approximate_theorem})\\
	\hline
	\end{tabular}
	  \begin{tablenotes}[para, online,flushleft]
	\footnotesize
$\mathcal{Q}$ is the set of all queries, $\mathcal{M}^*$ is the optimal cache, $N$ is the number of queries, $M$ is the size of cache, $l_{min}$is the minimum number of queries required to fill the cache as much as possible, $P(q)$ is the sampling probability of query $q$, $T$ is the time horizon, $\Delta_q, \Delta_q^c$ represent the complementary gap and the approximate gap between the cache containing $q$ and the optimal cache respectively. From the table, \cite{Berkeley} lacks an analysis of the problem-dependent form of regret, and when \texttt{Oracle} provides an exact solution, the regret bound of our algorithm outperforms that of \cite{Berkeley} the coefficient of $\sqrt{MN}$.
	\end{tablenotes}
			\end{threeparttable}
	}
\end{table*}

Table \ref{tab:result} summarizes our key findings and provides a detailed comparison with prior work, demonstrating the advantages of our approach in terms of both generalization and computational efficiency.

\section{Related work}
\textbf{Cache for LLMs:} In the field of LLM, cache technology can address the issue of repeated inquiries, thereby reducing the cost of the LLM marketplace \cite{frugalgpt}. \cite{xu2023sparksgptsedgeintelligence,zhang2023h2oheavyhitteroracleefficient} focus on content reuse during the inference phase, aimed to optimize the internal data flow within LLMs. \cite{bang-2023-gptcache} is recognized in the industry for its simplicity and flexibility, providing various options for integration with different LLMChat services. \cite{SCALM} leverage semantic understanding of queries to improve cache hit rates through a semantic-oriented caching approach.
Beyond these works, \cite{Berkeley} propose an online learning-based caching algorithm to minimize resource waste. While they briefly discuss the potential of extending their method to accommodate variable-sized queries, they do not provide a rigorous theoretical analysis or concrete solutions. Our work builds upon their approach by generalizing caching to variable-size constraints, addressing a critical gap in dynamic query management.

\textbf{Knapsack problem:} The Knapsack Problem \cite{knapsack} is a fundamental problem in optimization and computer science, frequently used in resource allocation and selection tasks. 
In our work, the knapsack problem serves as the foundation for our cache selection strategy, helping determine which queries should be stored or evicted efficiently. Specifically, we model our problem as a 0-1 Knapsack Problem and employ dynamic programming \cite{Dynamicprogramming} as an \texttt{Oracle} for updating our cache replacement policy.
Additionally, the Online Knapsack Problem \cite{10.1145/1367497.1367747} is a variant where decisions must be made without knowledge of future inputs, making it particularly relevant to online caching scenarios. Another relevant extension is the Minimum Knapsack Problem (MKP) \cite{minknapsackproblem,SANTINI2024385}, which differs from the traditional 0/1 Knapsack by seeking to minimize an objective while satisfying specific constraints, rather than maximizing it. These formulations inform our approach to adaptive cache selection under variable query sizes.

\textbf{Multi-armed bandit:} 
A natural extension of multi-armed bandit is the Combinatorial Multi-Armed Bandit (CMAB) \cite{CUCB,CMABT}, where a super arm consists of multiple base arms that are played simultaneously. In our work, we model each query as an arm, with its cost following an unknown distribution. Under this formulation, Upper Confidence Bound (UCB)-type methods provide a principled approach to managing uncertainty and making optimal caching decisions. From the perspective of the CMAB, we can regard the cache as a super arm. However, CMAB model cannot be directly adapted to our setting. On the one hand, feedback can only be obtained for queries that are not selected, which is the opposite of the traditional bandit feedback mechanism. On the other hand, in our online setting, each cache selection is restricted. Unlike CMAB, where super arms can be freely composed from all base arms, in our setting, the cache can only be chosen from the current cache and the query that has just arrived.

\section{Problem Setting}\label{section:setting}
\subsection{Caching with Variable Size Query}

We consider a finite set of queries $\displaystyle \mathcal{Q} =\{q_1,...,q_{N}\}$ containing $N$ distinct queries, each query $q \in \mathcal{Q}$ is associated with a size $L(q)\in \mathbb{N}_+$. When query $q$ is input to the LLM system, the LLM processes it and returns a corresponding result with size $A(q)\in\mathbb{N}_+$. We use $S(q)=L(q)+A(q)$ to denote the \textit{total query size} of $q$. Each distinct query has a sampling probability, which represents its frequency of occurrence. In each round, a query is received,  which is sampled from $\mathcal{Q}$ according to the fixed sample distribution. We use $P(q) \in (0,1]$ to denote the sampling probability that query $q$ be selected such that $\sum_{q\in \mathcal{Q}}P(q)=1$. Each time the LLM processes $q$, it will incur a random cost $C(q)$, which is supported on $[c_1,c_2]$ with $c_2>c_1>0$. The cost can be expressed as: 
\[
\textstyle
C(q) = C^*(q)  + \epsilon_q.
\]
where $\epsilon_q$ is a sub-Gaussian noise that captures the uncertainties in the cost, with $\mathbb{E}[\epsilon_q] = 0$. 
The expected true cost $C^*(q)=\mathbb{E}[C(q)]$ is the mean cost of processing query $q$. 
This model is aligned with \cite{Berkeley}.

\begin{remark}
Since the cache needs to store both the query for identification and the answer for direct response, we refer to the total query size as query size for simplicity.
\end{remark}
The cache stores both the query and its result content, and outputs the result of a matching query directly with zero cost. We use $r(q)$ to denote the result content return by LLM correspond to query $q$. We maintain a cache $\mathcal{M}$ for the LLM system, storing a small subset of queries with their corresponding results, i.e. we store a question-answer pair $(q,r(q))$ in cache, and the size of $(q,r(q))$ is $S(q)$. We use $q\in \mathcal{M}$ to denote $(q,r(q))$ has been stored in cache. The size of cache has a maximum size $M$, the total size of the stored queries must satisfy $\sum_{q \in \mathcal{M}} S(q) \leq M$. Let $\mathfrak{J}$ be the set of all possible sets that satisfy the size constraint
\[
\textstyle
\mathfrak{J} 
= \left\{\mathcal{M}|\sum\nolimits_{q\in \mathcal{M}}S(q)\leq M\right\}.
\]
To reduce overhead, we should place queries with high expected costs and high sampling frequencies into the cache. Following \cite{Berkeley}, we can define the expected cost of a given cache $\mathcal{M}$ is
\[
\textstyle
cost(\mathcal{M})=\sum\nolimits_{q\in \mathcal{Q}}C^*(q)P(q)\mathbb{I}\{q \notin \mathcal{M}\},
\]
where $\mathbb{I}$ is the indicator function. Note that $\mathbb{I}\{q\notin \mathcal{M}\}+\mathbb{I}\{q\in \mathcal{M}\}=1$. From another perspective, the cost we saved by cache hitting can be regarded as additional reward we have obtained. Then optimal cache can be define as 
\begin{align}
\textstyle
\mathcal{M}^*=argmax_{\mathcal{M}\in \mathfrak{J}}\sum_{q\in\mathcal{M}}C^*(q)P(q).
\end{align}
Previous works such as \cite{Berkeley} assume homogeneous query sizes, where $S(q)=1$ for all queries, reducing the problem to a top-$K$ selection problem. In contrast, we consider a heterogeneous setting, where queries have variable sizes $S(q)$, extending previous work.

\textbf{\texttt{Oracle}.} From the formula of optimal cache, we can model the optimal problem as a 0-1 knapsack problem where $C^*(q)P(q)$ represents the value of $q$ and $S(q)$ is its volume. The cache is treated as a knapsack with capacity of $M$, and the optimal cache is the optimal solution of a knapsack problem. Inspired by the CMAB model \cite{CUCB}, we assume that the agent has access to an offline \texttt{Oracle} to guide cache selection decisions. \texttt{Oracle} have an offline solver that can solve the combinatorial optimization problem. In the case described above, the solver employs an algorithm for solving the knapsack problem.

In some cases, the solution provided by the solver may be approximate, leading to approximate \texttt{Oracle}. When \texttt{Oracle} is approximate, we reformulate the problem that \texttt{Oracle} need to solve as a minimum knapsack problem (MKP), with the solver then acting as the algorithm for solving the MKP. 
Details of this approach is in 
Section \ref{sec:apx}.

\subsection{Online Learning}
\textbf{Online cache bandit.} The agent's objective is to choose the optimal cache without knowing the expected true cost and sample distribution. We consider a total number of $T\in \mathbb{N}_+$ learning rounds. In round $t$, agent receives a query $q_t$ which is sampled from $\mathcal{Q}$ with probability $P(q)$. The agent has a \textit{current cache} $\mathcal{M}_t \in \mathfrak{J}$ in round $t$. $\mathcal{M}_t$ is selected based on the history feedback information from previous rounds. After $q_t$ arrive, the agent will check the current cache $\mathcal{M}_t$. 
If $q_t$ is found in the cache, i.e. $q_t \in \mathcal{M}_t$, we say that query \textit{ hits} the cache. In this case, the result of $q_t$ is directly returned without further processing by the LLM. The cost of processing this query is $0$ and will save a potential cost $C(q_t)$, which is unobserved to the agent. Therefore, the feedback received during a cache hit would be: the arrival query $q_t$ without cost. 
If query $q_t$ \textit{miss} the cache, i.e. $q_t \notin \mathcal{M}_t$, the system processes the query, incurring a cost $C(q_t)$, returns the result, and delivers the feedback to the agent. Feedback received during a missing cache would be: the arrival query $q_t$ with cost $C(q_t)$ and receiving the content of the result $r(q_t)$ from LLM and its size $S(q_t)$.

The agent can then choose to update the cache based on history feedback information or keep the current cache unchanged. In the same time, the algorithm will decide, based on historical information, whether to use \texttt{Oracle}. If \texttt{Oracle} is called, the cache will be updated according to the output provided by \texttt{Oracle}.
\begin{remark}  
\label{remark:reversebandit}
As noted in \cite{Berkeley}, our setting differs from traditional bandit problems because feedback may not be observed in every round. Specifically, when a query hits the cache, no cost feedback is received.
\end{remark}
\begin{remark} 
\label{remark:online}
In the online mode, the agent's updating of the cache in each round are constrained. At each time step, the cache can only be updated using the currently arriving query and the queries already stored in the cache, i.e. $\mathcal{M}_{t+1}\subseteq\mathcal{M}_{t}\cup \{q_t\}$. This is because that we discard the results $r(q)$ of queries that are removed from the cache during updates (but we retain information such as the estimated sample probability, cost, and size of the query), only the query with $r(q)$ can be putted in the cache. If a previously dropped query reappears, the $r(q)$ need to be recomputed by LLM.
\end{remark}
\textbf{Learning objective.}
Our goal is to minimize the total cost with time horizon $T$. By applying the indicator function to the regret in \cite{Berkeley} for an equivalent transformation, we define the regret as:
\[\textstyle
Reg(T)=\mathbb{E} \left[\sum_{t=1}^{T}C(q_{t})\left (\mathbb{I}\{q_{t}\in \mathcal{M}^*\}-\mathbb{I}\{q_{t}\in \mathcal{M}_t\}  \right ) \right],
\]
where $\mathcal{M}^*=argmax_{\mathcal{M}\in \mathfrak{J}}\sum_{q\in\mathcal{M}}C^*(q)P(q)$.

\section{Variable Size Online Cache Bandit}
\subsection{Algorithm with Exact \texttt{Oracle}}
Our algorithm is presented in Alg.\ref{alg:main}. When we have an exact solver, $Oracle$ in line 25 refers to Alg.\ref{alg:exactOracle}.

\textbf{Query record.} In the online setting, the agent does not know which queries exist at the beginning. Therefore, the agent needs to maintain a set to record the queries it has encountered. Only after the agent has seen a query will it have information about that query and estimate its cost and sampling probability. Subsequently, when invoking \texttt{Oracle} or updating the cache, the agent will take this query into account. We use a set $\mathcal{Q}_t$ (line 2) in our algorithm to represent this record set, a query $q\in \mathcal{Q}_t$ means that the estimation and information of this query (excluding $r(q)$) have been recorded by algorithm. Clearly, when all the queries in $\mathcal{Q}$ have arrived at least once, we have $\mathcal{Q}_t = \mathcal{Q}$.

\textbf{\texttt{Oracle} guide:} For heterogeneous query sizes, the previously straightforward single-query replacement strategy is no longer applicable. This is because, with variable size query, we also need to consider whether the length of the query allows for a valid replacement during cache replacement. A query might not fit into the cache due to size constraints, causing the replacement to fail. In some cases, this could result in a scenario where no replacements occur, ultimately trapping the system. Inspired by the \texttt{Oracle} in the CMAB model \cite{CUCB}, we assume that the algorithm can use an \texttt{Oracle} that accepts global query information and returns an optimal cache selection under the given constraints. As we mentioned above, solving for the optimal cache can be viewed as a knapsack problem. Specifically, \texttt{Oracle} is denoted by 
\[Oracle(\mathcal{Q}',\{C^*(q)\}_{q \in \mathcal{Q}'},\{P(q)\}_{q \in \mathcal{Q}'},\{S(q)\}_{q \in \mathcal{Q}'},M).\]
It takes the cost, sample probability and query size of all queries in a given set $\mathcal{Q}'$ as input and output a set $\mathcal{M}$ that yields the solution of knapsack problem with volume constraint of $M$. 

We assume that \texttt{Oracle} has a knapsack problem solver (e.g. dynamic programming) which can output the exact optimal knapsack solution. We denote this solver as $Knap(\bm{w},\bm{v},m)$. By inputting the item values $\bm{w}$, volumes $\bm{v}$, and the knapsack capacity $m$ into $Knap$, it will return the optimal knapsack solution. We encapsulate this solver as our \texttt{Oracle} in Alg.\ref{alg:exactOracle}. \texttt{Oracle} accepts the estimated values of query costs and sampling probabilities, multiplies them together to compute the value, and then inputs this value into $Knap$. Compared to the typical single-query replacement, \texttt{Oracle} can see global information, allowing it to provide global guidance on cache replacement. 

\textbf{Lower confidence bound:} Similar to \cite{Berkeley}, we use a lower-confidence-bound (LCB) for estimating the cost of each query. This LCB approach is akin to adding a penalty to the estimation of queries that remain in the cache for a long time, making them more likely to be replaced, similar to an enforcement of exploration. We use $C_{t}(q)$ to denote the cost of query $q$ that the agent receives in round $t$. In algorithm, we use $T^{(c)}_{t}(q)=\sum_{s=1}^{t}\mathbb{I}\{q_s\notin \mathcal{M}_s, q_s=q\}$ to represent the number of cache misses for query $q$ up to the $t$-th round, where $\mathbb{I}\{q_s\notin \mathcal{M}_s, q_s=q\}$ denote query
$q$ arrive in round $s$ and misses cache. Let $\widetilde{C}_{t}(q)=\sum_{s=1}^{t}\mathbb{I}\{q_s\notin \mathcal{M}_s, q_s=q\}C_{t}(q)$ to represent the accumulated cost information of query $q$ up to the $t$-th round. The lower-confidence-bound based estimator for cost is:
\begin{equation}
\label{eq:cost_estimate}
\textstyle
\widehat{C}_{t}(q) 
{=} 
\max\left\{0,\frac{\widetilde{C}_{t}(q)}{T^{(c)}_{t}(q)}-(c_2-c_1)\sqrt{\frac{\ln{(8TN/\delta)}}{2T^{(c)}_{t}(q)}}\right\}.
\end{equation}
\textbf{Variance confidence bound:} Different from \cite{Berkeley}, for the estimation of the sampling probability $P(q)$, we also used an LCB-type estimate rather than a empirical estimate. Additionally, the confidence bound for $\widehat{P}_t(q)$ is derived from the variance bound in \cite{Bernstein}. According to the work in \cite{Bernstein}, the variance estimate $V_t(q)$ of the sampling probability of a query can be obtained in the following form:
\begin{align}
\nonumber
V_t(q)=
&
\textstyle
\frac{1}{t}\sum^{t}_{s=1}\left(\mathbb{I}\{q_s=q\}-\frac{\sum_{i=1}^{t}\mathbb{I}\{q_i=q\}}{t} \right)^2.
\end{align}
Let $T^{(q)}_{t}(q)=\sum_{s=1}^{t}\mathbb{I}\{q_s=q\}$, the estimator for sample probability in round $t$ is:
\begin{align}
\label{eq:P_estimate1}
&
\textstyle
\widehat{P}_{t}(q)=\max 
\left\{ 
0,\frac{T^{(q)}_{t}(q)}{t}-LVCB_t(q,\delta) 
\right\},\\
\label{eq:P_estimate2}
&
\textstyle
LVCB_t(q,\delta) 
{=} \sqrt{\frac{3V_t(q)\ln{(16TN/\delta)}}{t}} {+} 
\frac{5\ln{(16TN/\delta)}}{t}.
\end{align}
\textbf{Recommend and wait strategy:} In online model, the algorithm does not store the result content corresponding to all queries. When the cache is updated, the results content $r(q)$ of the queries that are removed from the cache will be cleared. Because cache need to store $(q,r(q))$, cache selection can only access the queries currently in the cache and any newly arriving queries. As mentioned in Remark\ref{remark:online}, the queries chosen for cache updates are not arbitrary. Although \texttt{Oracle} can output an optimal solution for the global set of queries, unlike the setting in CMAB \cite{CUCB}, our cache cannot be directly updated to the selection recommended by \texttt{Oracle} due to the online constraint. For example, a query in \texttt{Oracle}'s output indicates that the agent should add this query to the cache, but if this query was replaced from the cache during a previous update, the agent will not have $r(q)$ for this query in the current round. In this case, the cache must wait until the query arrives again and the $r(q)$ is re-obtained before it can be added to the cache. Under this setting, we involve a \textit{Recommended cache} denoted by $\widehat{\mathcal{M}}_t$ to temporarily store the results output by our \texttt{Oracle} at the current time. When \texttt{Oracle} is invoked, the policy does not update the cache immediately. Instead, it first receives a recommended cache $\widehat{\mathcal{M}}_{t}$, and the cache will retain only the portion that matches the recommended cache, and the differing parts will be removed to free up space for subsequent updates (line 25-26). In subsequent rounds, the algorithm updates the current cache based on recommended cache. In the following rounds, if a query that belongs to the recommended cache but is not currently in the cache arrives, it will miss the current cache and be processed by the LLM to obtain its response. The query will then be added to the current cache. Otherwise, the cache will not be updated unless there is still available space in the recommended cache. This process continues until the next recommended cache update occurs (line 18-22).

\textbf{Full cache:} The recommended cache may not be full until all queries have been seen at least once (i.e. $\mathcal{Q}_t = \mathcal{Q}$). This means that there may still be unused space in recommended cache. To avoid wasting space, we check whether there is still available space in the recommended cache (line 20). If so, we place additional queries into the cache, regardless of whether they belong to the recommended cache (line 18).

\textbf{Accumulation-based \texttt{Oracle} invoking policy:} The calling of \texttt{Oracle} can gradually update the current cache to the optimal solution, but it also introduces additional computational overhead for the entire system. Especially when solving such knapsack problems, the cost of calling \texttt{Oracle} is substantial. Therefore, we need to minimize the number of \texttt{Oracle} calls as much as possible, while still minimizing the impact on the cache updates. In our algorithm, we use an accumulation-based strategy (line 23) to balance this trade-off. $T^{(r)}_t(q)$ represents the number of cache misses for query $q$ at the time of the most recent \texttt{Oracle} call before time $t$. Accumulation means that an \texttt{Oracle} call occurs when the number of cache misses $T^{(c)}_t(q)$ in current for a query exceeds $1 + \alpha$ times $T^{(r)}_t(q)$, where $\alpha>0$ is a predetermined constant. This is equivalent to calling \texttt{Oracle} only when the query has accumulated enough information about its cost, making the estimation of its cost more accurate. This approach is inspired by the asynchronous update strategy in federated bandit methods \cite{federatedbandit}. Since our algorithm needs to estimate both the cost and the sampling probability of queries, our update strategy also maintains a timer $T_t$ to represent the time of the most recent \texttt{Oracle} call before $t$. When the number of rounds $t$ reaches $(1 + \alpha)T_t$, we also perform an \texttt{Oracle} call. The role of the timer $T_t$ is to accumulate information about the query's sampling probability, so that when the estimate of the sampling probability becomes more accurate, \texttt{Oracle} can be called in a timely manner. This accumulation-based \texttt{Oracle} invoking policy ensures that \texttt{Oracle} is called in a timely manner, once the estimates become relatively accurate, while also keeping the number of \texttt{Oracle} calls logarithmic in scale.

\begin{algorithm}[!t]
    \caption{Variable Size Online Cache Bandits (VSOCB)} 
    \label{alg:main}
    \begin{algorithmic}[1]
    \REQUIRE $\mathcal{Q}_0=\phi$. $T_0=0$. For all $q \in \mathcal{Q}$: $T_{0}^{(r)}(q)=T^{(q)}_{0}(q)=T^{(c)}_{0}(q)=0$, $\widehat{P}_{0}(q)=V_0(q)=0$, $\widetilde{C}_{0}(q)=\widehat{C}_{0}(q)=0$, $S(q)=0$. $\mathcal{M}_{0}=\widehat{\mathcal{M}}_{0}=\phi$, $\alpha >0$.
    \ENSURE ~~\\ 
    \FOR{round $t=1,...,T$}
	\STATE $q_{t}$ was received with size $L(q_t)$, $\mathcal{Q}_{t}=\mathcal{Q}_{t-1}\cup \{q_t\}$
	\STATE $T^{(q)}_{t}(q_t)=T^{(q)}_{t-1}(q_t)+1$
	\IF{$q_{t}\in \mathcal{M}_{t-1}$}
	\STATE Output the response from cache with 0 cost
	\STATE $\widetilde{C}_{t}(q_t)=\widetilde{C}_{t-1}(q_t)$, $T^{(c)}_{t}(q_t)=T^{(c)}_{t-1}(q_t)$
	\ELSE
	\STATE Input $q_{t}$ to LLM and receive $r(q_t), A(q_{t}), C_{t}(q_{t})$.
	\STATE $S(q_t)=L(q_t)+A(q_t)$ //record query size
	\STATE $\widetilde{C}_{t}(q_t)=\widetilde{C}_{t-1}(q_t)+C_{t}(q_{t})$
        \STATE $T^{(c)}_{t}(q_t)=T^{(c)}_{t-1}(q_t)+1$
	\ENDIF
	\FOR{all $q \ne q_t$}
	\STATE $T^{(q)}_{t}(q)=T^{(q)}_{t-1}(q)$, $T^{(r)}_t(q)=T^{(r)}_{t-1}(q)$
	\STATE $\widetilde{C}_{t}(q)=\widetilde{C}_{t-1}(q)$, $T^{(c)}_{t}(q)=T^{(c)}_{t-1}(q)$
	\ENDFOR
        \STATE compute $\widehat{P}_{t}(q), \widehat{C}_{t}(q)$ with Eq. (\ref{eq:cost_estimate}), (\ref{eq:P_estimate1}), (\ref{eq:P_estimate2}), $\forall q \in \mathcal{Q}_t$
	\IF{$q_t\in \widehat{\mathcal{M}}_{t-1}$ or $S(q_t)\leq M-\sum_{q\in \widehat{\mathcal{M}}_{t-1}}S(q)$} \label{tec:balance}
	\STATE $\mathcal{M}_{t}=\mathcal{M}_{t-1}\cup \{q_t\}$, $\widehat{\mathcal{M}}_{t}=\widehat{\mathcal{M}}_{t-1}\cup \{q_t\}$ 
        \ELSE
	\STATE $\mathcal{M}_{t}=\mathcal{M}_{t-1}$, $\widehat{\mathcal{M}}_{t}=\widehat{\mathcal{M}}_{t-1}$
	\ENDIF
        \label{line:acc}
	\IF{$T^{(c)}_t(q_t)\geq (1+\alpha)T^{(r)}_{t-1}(q_t)$ or $t\geq (1+\alpha)T_{t-1}$}
	\STATE $T^{(r)}_t(q_t)=T^{(c)}_t(q_t)$, $T_t=t$
        \label{line:Oracle}
	\STATE $\widehat{\mathcal{M}}_{t}=Oracle(\mathcal{Q}_t,\widehat{C}_{t},\widehat{P}_{t},S,M)$
    \STATE $\mathcal{M}_{t}=\widehat{\mathcal{M}}_{t}\cap \mathcal{M}_{t-1}$ 
	\ELSE
	\STATE $T^{(r)}_{t}(q_t)=T^{(r)}_{t-1}(q_t)$, $T_t=T_{t-1}$, $\widehat{\mathcal{M}}_{t}=\widehat{\mathcal{M}}_{t-1}$
	\ENDIF
    \ENDFOR 
    \end{algorithmic}
    \end{algorithm}

\begin{algorithm}[!t]
    \caption{Exact \texttt{Oracle}} 
    \label{alg:exactOracle} 
    \begin{algorithmic}[1]
    \STATE Input $\mathcal{Q}'$, $\{C'(q)\}_{q\in \mathcal{Q}'}$, $\{P'(q)\}_{q\in \mathcal{Q}'}$, $\{S'(q)\}_{q\in \mathcal{Q}'}$, $M$
	\STATE $w(q)=P'(q)C'(q), \forall q \in \mathcal{Q}'$
	\STATE $\mathcal{M} \leftarrow Knap(\{w(q)\}_{q\in \mathcal{Q}'},\{S'(q)\}_{q\in \mathcal{Q}'},M)$
    \STATE Output $\mathcal{M}$
    \end{algorithmic}
    \end{algorithm}
\subsection{Exact Solution Bound}
As mentioned in the algorithm above regarding the full cache, we always aim to fill the cache as much as possible, meaning that no additional query can be added to the cache. Under this requirement, similar to CUCB \cite{CUCB}, we define the valid set of cache.
\begin{definition}
(Valid set): We define the valid set as 
\[
\textstyle
\mathfrak{V} {=} 
\!\left\{\! \mathcal{M}|\mathcal{M} {\subseteq} \mathcal{Q}, 
M {-} \min_{q'\notin \mathcal{M}} \!S(q') {<} \sum_{q\in \mathcal{M}} \!S(q) {\leq} M 
\!\right\}\!.
\]
We define $l_{max}=\max_{\mathcal{M}\in \mathfrak{V}} |\mathcal{M}|$, $l_{min}=\min_{\mathcal{M}\in \mathfrak{V}} |\mathcal{M}|$ to represent the maximum and minimum number of queries that can be included in the valid set, respectively.
\end{definition}
\begin{definition}
(Complementary Gap): For each $\mathcal{M}\in \mathfrak{V}$, we define the gap $\Delta_{\mathcal{M}}=\sum_{q\in \mathcal{M}^{*}}P(q)C^*(q)-\sum_{q\in \mathcal{M}}P(q)C^*(q)$. For each query $q$, we define $\Delta_{q}=\min_{\mathcal{M}\in \mathfrak{V}:\mathcal{M}\ne \mathcal{M}^*,q\notin \mathcal{M}}\Delta_{\mathcal{M}}$. Similar to the definition 2 in \cite{CMABT}, if there is no cache selection $\mathcal{M}$ such that $q \in \mathcal{M}$ and $\mathcal{M}\ne \mathcal{M}^*$, we define $\Delta_{q}=+\infty$. Then we define $\Delta_{min}=\min_{q \in \mathcal{Q}}\Delta_{q}=\min_{\mathcal{M}\in \mathfrak{V}}\Delta_{\mathcal{M}}$.
\end{definition}
\begin{remark}
We define the gap of a query as the difference between the non-optimal cache that does not include this query and the optimal cache. This gap of query measures the impact when a cache does not select this query. Detailed discussion is in Appendix.
\end{remark}
\begin{theorem}
\label{thm:main_theorem}

Setting $\delta=1/T$, 
(1) if $\Delta_{min}>0$, the problem-dependent regret bound for VSOCB (Alg.\ref{alg:main}) with exact \texttt{Oracle}(Alg.\ref{alg:exactOracle}) holds:
\begin{align}
\nonumber  
& 
\textstyle
Reg(T)\leq \widetilde{O}\left(\sum_{q\in\mathcal{M}^*}\frac{P(q)N l(M,\mathcal{Q})}{\Delta_q}\right),
\end{align}
(2) and problem-independent regret bound:
\begin{align}
\nonumber
&Reg(T)\leq \widetilde{O}\left(\sqrt{N l(M,\mathcal{Q})T}\right),
\end{align}
where $l(M,\mathcal{Q})=\min\{l_{max},N-l_{min}\}$.
(3) and the total number of \texttt{Oracle} calls is $O(N\log{T})$.
\end{theorem}
This proof is deferred to Appendix. In the proof, we adopted an analysis method similar to \cite{CMABT} and combined it with the \texttt{Oracle} invoking policy proposed in our algorithm. We show that the accumulation-based \texttt{Oracle} invoking policy associates the duration for which the cache remains unchanged. Our conclusion suggests that both very small and very large caches result in relatively low regret. This is intuitive because, if the cache is very small and can only hold one query, we will have $l(M,\mathcal{Q})=1$, the optimal cache can be easily identified. In this case, the algorithm can quickly find the best cache. If the cache is large enough to hold all queries, then $l(M,\mathcal{Q})=0$, and the cache can store all queries, which eliminates regret altogether.

We demonstrate that our Online Variable Size Cache Bandits achieves the regret of $\widetilde{O}(\sqrt{MNT})$ (becasue $l(M,\mathcal{Q})\leq M$), which represents an improvement over Caching in Online Learning in \cite{Berkeley} with the coefficient of $\sqrt{MN}$, where the conclusion of \cite{Berkeley} was $\widetilde{O}(MN\sqrt{T})$. On the other hand, our model is an extension of \cite{Berkeley}, which can be regarded as a special case of our model. In this case, \cite{Berkeley} proposed a lower bound of $\Omega(\sqrt{T})$. The problem-independent regret established in Theorem \ref{thm:main_theorem} conforms to $\widetilde{O}(\sqrt{T})$,  indicating that our work provides an optimal solution in relation to $T$.
 
Although our algorithm requires an additional \texttt{Oracle} to assist in computation, our proof shows that the number of calling \texttt{Oracle} is $O(N\log{T})$, which is logarithmic. This indicates that the additional computational cost caused by using \texttt{Oracle} is not significant, and it well balances the trade-off between computational cost and regret.
\section{VSOCB with Approximate \texttt{Oracle}}
\label{sec:apx}
\subsection{Approximate \texttt{Oracle}}
When the solver can only output an approximate solution, we require the solver to output the complement of the cache. In this case, the optimization problem for solver transforms into a min-knapsack problem \cite{minknapsackproblem}:
\begin{align}
\nonumber
& 
\textstyle
\min_{\mathcal{M}}{\sum_{q\in \mathcal{M}^c}P(q)C^*(q)} \\
\nonumber
&s.t. 
\textstyle
\sum_{q\in \mathcal{M}^c}S(q)\geq \sum_{q\in \mathcal{Q}}S(q)-M,
\end{align}
where $q\in \mathcal{M}^{c}$ means that $q\notin \mathcal{M}$. We use $knap_{min}$ to denote min-knapsack problem solver and it output the a solution of MKP. Clearly, the optimal solution set for this MKP is the complement of the optimal solution set for the original knapsack problem. We use $\mathcal{M}^{*c}=\mathcal{Q}\setminus\mathcal{M}^{*}$ to denote the optimal complementary of cache. 

\begin{definition}
($\beta$-approximation solver): A minimum knapsack problem solver is $\beta$-approximation solver mean that: the output $\mathcal{M}^c$ of solver under input set $\mathcal{Q}'$ and input value $\{P'(q)C'(q)\}_{q\in \mathcal{Q}'}$ satisfy $\sum_{q\in\mathcal{M}^c}P'(q)C'(q)<(1+\beta)\sum_{q\in\mathcal{M}^{*c}}P'(q)C'(q)$
\end{definition}

Since the minimum knapsack solver outputs the complement of the knapsack problem solution, we take the complement of the solver's output to obtain the recommended cache, \cref{alg:apxOracle} is our approximate \texttt{Oracle}. Replacing the $Oracle$ in Alg.\ref{alg:main} with Alg.\ref{alg:apxOracle} is the approximate version of VSOCB.

Remarking that we define the approximate algorithm on the minimum knapsack problem instead of directly defining an approximate algorithm on the original knapsack problem. This mainly due to our theoretical analysis, when \texttt{Oracle} outputs an approximate solution, it cannot directly compare the optimal cache with the recommended cache in the same way as an \texttt{Oracle} that provides an exact solution. We relax the regret analysis to focus on the complement, using this relaxation to address this challenge.
\begin{algorithm}[t]
    \caption{Approximate \texttt{Oracle}} 
    \label{alg:apxOracle} 
    \begin{algorithmic}[1] 
    \STATE Input $\mathcal{Q}'$, $\{C'(q)\}_{q\in \mathcal{Q}'}$, $\{P'(q)\}_{q\in \mathcal{Q}'}$, $\{S'(q)\}_{q\in \mathcal{Q}'}$, $M$
	\STATE $w(q)=P'(q)C'(q), \forall q \in \mathcal{Q}'$
	\STATE $\mathcal{M}^c\leftarrow Knap_{min}(\{w(q)\}_{q\in \mathcal{Q}'},\{S'(q)\}_{q\in \mathcal{Q}'},M)$
    \STATE Output $\mathcal{Q}'\setminus \mathcal{M}^c$
    \end{algorithmic}
    \end{algorithm}

\subsection{Approximate Solution Bound}
Since the output in the case of an approximate solver is the complement of the cache, we need to redefine some of the previous definitions. First, following by CMAB \cite{CUCB},we redefine the regret of our problem. 
\begin{definition}
($\beta$-approximation regret): For a $\beta$-approximation solver, the $\beta$-approximation regret is:
\begin{align}
\nonumber
\textstyle
Reg_{\beta}(T) =\sum_{t=1}^{T}\mathbb{E}\left[\sum_{q\in \mathcal{M}_t^{c}}P(q)C^*(q)-(1+\beta)\sum_{q\in \mathcal{M}^{*c}}P(q)C^*(q)\right].
\end{align}
\end{definition}
\begin{definition}
(Approximation gap): For each $\mathcal{M}\in \mathfrak{V}$ and $\mathcal{M}^c=\mathcal{Q}\setminus \mathcal{M}$, we define the gap $\Delta_{\mathcal{M}}^c=\sum_{q\in \mathcal{M}_{t}^{c}}P(q)C^*(q)-(1+\beta)\sum_{q\in \mathcal{M}^{*c}}P(q)C^*(q)$. For each query $q$, we define $\Delta_{q}^c=\min_{\mathcal{M}\in \mathfrak{V}:\mathcal{M}\ne \mathcal{M}^{*},q\notin \mathcal{M}}\Delta_{\mathcal{M}}^c$. If there is no $\mathcal{M}^c$ such that $q \in \mathcal{M}^c$ and $\mathcal{M}^c\ne \mathcal{M}^{*c}$, we define $\Delta_{q}^c=+\infty$. We define $\Delta_{min}^c=\min_{q \in \mathcal{Q}}\Delta_{q}^c=\min_{\mathcal{M}\in \mathfrak{V}}\Delta_{\mathcal{M}}^c$.
\end{definition}
\begin{theorem}
\label{thm:approximate_theorem}
Setting $\delta=1/T$, (1) if $\Delta_{min}>0$, the problem-dependent regret bound for VSOCB (Alg.\ref{alg:main}) with approximate \texttt{Oracle} (Alg.\ref{alg:apxOracle}) holds:
\begin{align}
\nonumber
& 
\textstyle
Reg_{\beta}(T)\leq \widetilde{O}\left(\sum_{q\in\mathcal{Q}}\frac{P(q)N(N-l_{min})}{\Delta_q^c}\right),
\end{align}
(2) our problem-independent regret bound:
\begin{align}
\nonumber
&Reg_{\beta}(T)\leq \widetilde{O}\left(\sqrt{N(N-l_{min})T}\right),
\end{align}
(3) and the total number of \texttt{Oracle} calls is $O(N\log{T})$.
\end{theorem}
Our result under the approximate \texttt{Oracle} is less favorable than that under the exact \texttt{Oracle}. This is because, in the case of an approximate solution, it is difficult to directly apply the properties of the exact solution. Consequently, we relax the regret analysis to account for the gap between the optimal cache under the approximate solution and the cache selected by the algorithm.
\section{Experiment}
\subsection{Simulation Dataset}
We compare performance of our algorithms and the baseline using cumulative regret in online learning with a simulated dataset. We use the method described in the appendix of \cite{Berkeley} which replaces one query with the smallest expected cost per-size to solve variable size case as the baseline. We consider 100 distinct queries and set the cache size to be 60. The expected of cost for each query is sample from a uniform distribution support on $[1,2]$. We repeat the simulation 10 times and plot the mean in the figure. The cost for each query in each round is sampled from a normal distribution with a mean equal to the expected value generated above. We employ a dynamic programming algorithm as our solver in \texttt{Oracle}. We use an offline version to compare, denoted as \textbf{offline}. In offline, we assume that the algorithm can update the cache arbitrarily based on \texttt{Oracle}'s output, and \texttt{Oracle} can be called in every round without considering computational overhead. 

\begin{figure}[h]
\vspace{-0.12 in}
	\centering
	\includegraphics[width=0.4\linewidth]{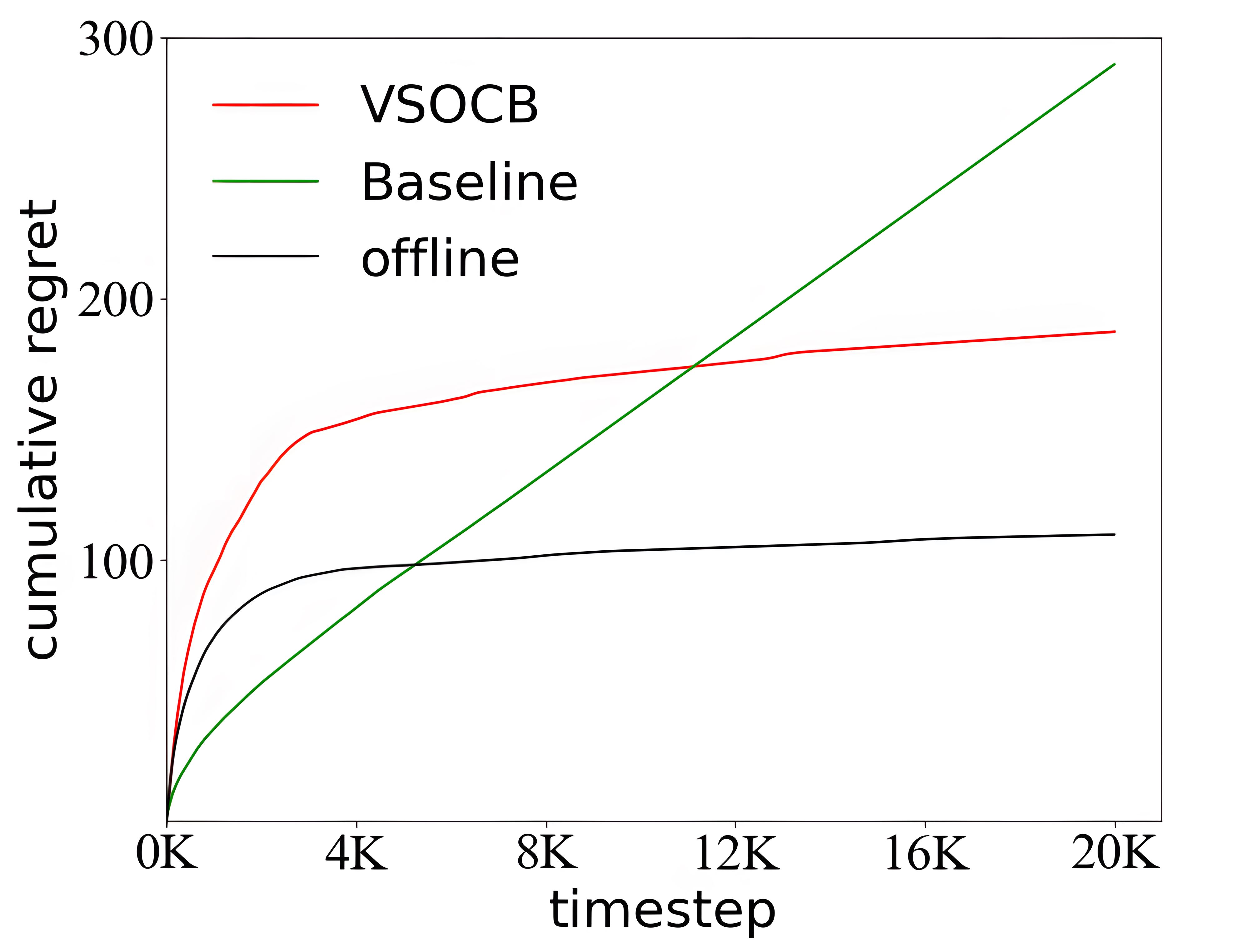}
    \vspace{-0.1in}
	\caption{Synthetic dataset with 20000 rounds}
    \vspace{-0.1in}
	\label{fig:syndata1}
\end{figure}

As shown in Fig.\ref{fig:syndata1}, our algorithm outperforms the baseline. This improvement can be attributed to the fact that the method in \cite{Berkeley} provides an approximate solution to the knapsack problem \cite{10.1145/1367497.1367747}, which poses a risk of falling into sub-optimal solutions. In contrast, our algorithm circumvents this issue by utilizing \texttt{Oracle} for the knapsack problem. We observe that VSOCB performs worse than offline. This discrepancy is due to the fact that offline algorithm can call \texttt{Oracle} in each round, allowing it to converge faster. Although offline algorithm exhibits lower regret, it requires substantial additional memory and computational resources, which may be unrealistic in real-world scenarios. Conversely, our VSOCB algorithm avoiding these significant additional costs to some extent.
\subsection{Real Dataset}
\begin{figure}[h]   
\vspace{-0.1in}
\centering 
\includegraphics[width=0.4\linewidth]{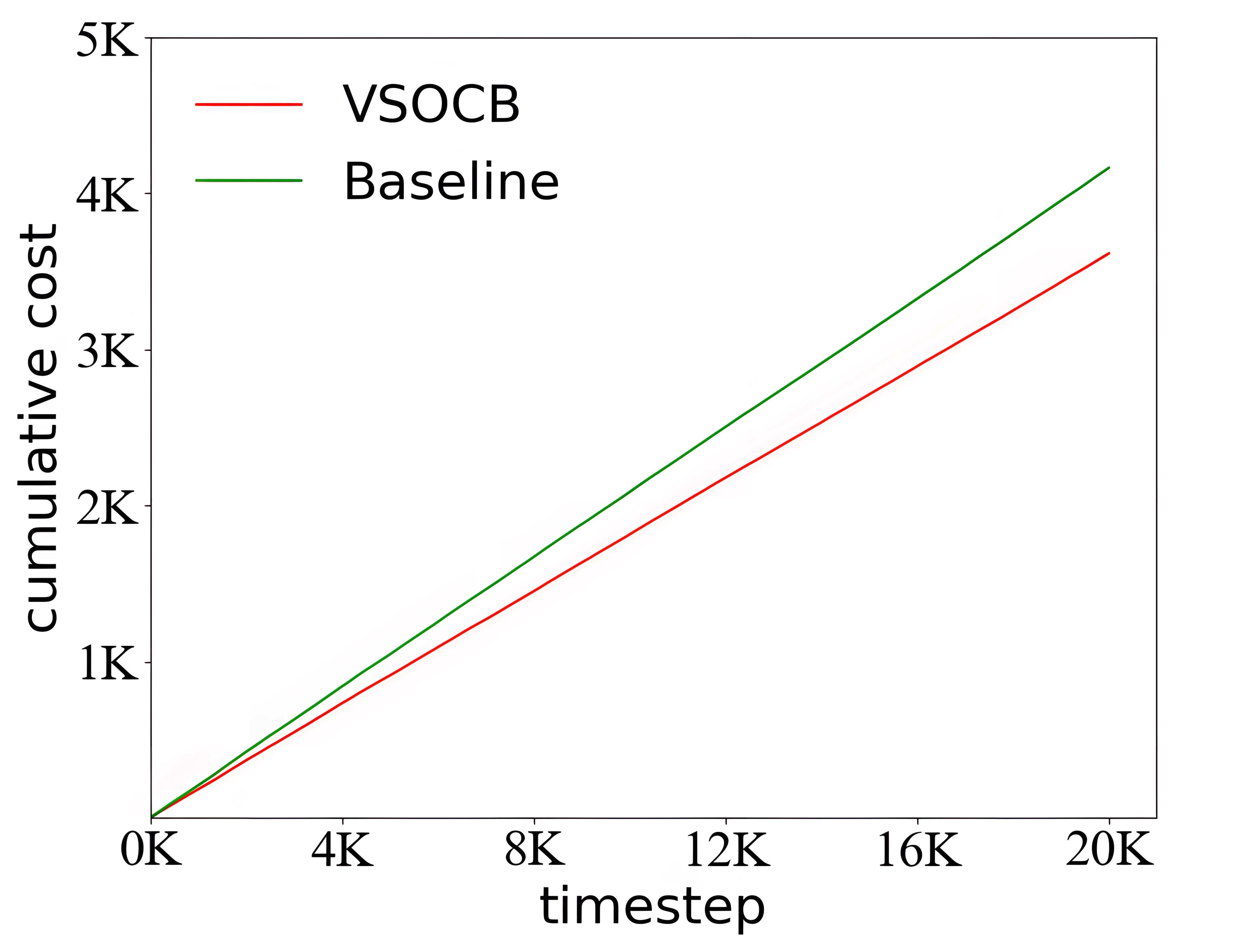}
	\caption{Average cumulative cost over 5 repeat runs on Real dataset with 20000 rounds}
	\label{fig:realdata1}
\end{figure}

In line with \cite{Berkeley}, we evaluate our algorithms using the OpenAssistant \cite{OpenAssistant} dataset for the chat assistant task. For this task, we employ the FastChat-T5-3B \cite{FastChat} model to implement our online algorithm, utilizing inference latency as the cost metric. We run our algorithm with 100 distinct queries in the online setting over a total of 20000 rounds, cache size is 100. As shown in Tab.\ref{tab:realdata1} and Fig. \ref{fig:realdata1}, our method reduced the cost by $13\%$ compared to the baseline, with the result averaged over 5 repetitions. After a sufficient number of online learning steps, \texttt{Oracle} accurately gets the costs and frequencies of each query, enabling VSOCB to outperform the baseline.

\begin{table}[h]
\centering
\begin{tabular}{llll}
\multicolumn{1}{c}{\bf Algorithm}  &\multicolumn{1}{c}{\bf Total cost} 
\\ \hline
 \cite{Berkeley}&  4162.8 \\ \hline
 VSOCB&  3614.2\\ \hline
\end{tabular}
\caption{Average cumulative cost on real dataset with 20000 rounds}
\label{tab:realdata1}
\end{table}
\section{Conclusion}
In this paper, we consider caching query in LLM system with heterogeneity size. We propose VSOCB for this problem with rigorous theoretical analysis. We prove that the regret bound of our algorithm is better than that of similar works, and we have also provided the problem-dependent regret bound form which is absent in previous work.

We acknowledge that there is still room for improvement in this work. Our lower bound is directly derived from \cite{Berkeley}, which only provides an analysis with respect to $T$, lacking considerations for other parameters such as $M,N$. On the other hand, due to the complexity of the technical analysis, we are limited to using approximate algorithms for the minimum knapsack problem. This results in a less favorable bound for the approximate solution compared to that of the exact solution.

\bibliographystyle{unsrt}  
\bibliography{arixiv_references}

\end{document}